\documentclass[letterpaper, 10 pt, conference]{ieeeconf}
\IEEEoverridecommandlockouts    
\usepackage{subcaption}
\usepackage{tabularx,booktabs}

\usepackage{graphics}           
\usepackage{times}              
\usepackage{amsmath}            
\usepackage{amssymb}            
\usepackage{graphicx}
\usepackage{algorithm}
\usepackage[noend]{algpseudocode}
\usepackage{booktabs}
\usepackage{color}
\definecolor{instructioncolor}{rgb}{.5,.5,.5}

\usepackage[font=small]{caption}

\def\secref#1{Sec.~\ref{#1}}
\def\figref#1{Fig.~\ref{#1}}
\def\tabref#1{Tab.~\ref{#1}}
\def\eqref#1{Eq.~(\ref{#1})}

\makeatletter
\usepackage{xspace}
\DeclareRobustCommand\onedot{\futurelet\@let@token\@onedot}
\def\@onedot{\ifx\@let@token.\else.\null\fi\xspace}

\def\etal{{et al}\onedot}
\makeatother

\def\etalcite#1{\etal~\cite{#1}}

\usepackage{array}
\newcolumntype{L}[1]{>{\raggedright\let\newline\\\arraybackslash\hspace{0pt}}m{#1}}
\newcolumntype{C}[1]{>{\centering\let\newline\\\arraybackslash\hspace{0pt}}m{#1}}
\newcolumntype{R}[1]{>{\raggedleft\let\newline\\\arraybackslash\hspace{0pt}}m{#1}}

\newcommand{\abs}[1]{|#1|}

\renewcommand{\b}[1]{\mbox{\boldmath$#1$}}

\newcommand{\q}[1]{{{\bf #1}}}

\newcommand{\mq}[1]{{\mbox{{\sffamily{#1}}}}}

\title{\LARGE \bf Constructing Metric-Semantic Maps using Floor Plan Priors \\ for Long-Term Indoor Localization }

\author{Nicky Zimmerman$^{\star}$ \and Matteo Sodano$^{\star}$ \and Elias Marks \and Jens Behley \and Cyrill Stachniss
  \thanks{$^\star$ Authors contributed equally to this work.}
  \thanks{All authors are with the University of Bonn, Germany. Cyrill Stachniss is additionally with the Department of Engineering Science at the University of Oxford, UK, and with the Lamarr Institute for Machine Learning and Artificial Intelligence, Germany. This work has been funded 
  by the European Union’s Horizon 2020 research and innovation programme under grant agreement No~101017008~(Harmony).
  }%
} 

\begin{document}  
\maketitle
\thispagestyle{empty}
\pagestyle{empty}

\begin{abstract}
  %
  Object-based maps are relevant for scene understanding since they integrate geometric and semantic information of the environment, allowing autonomous robots to robustly localize and interact with on objects. 
  In this paper, we address the task of constructing a metric-semantic map for the purpose of long-term object-based localization. 
  We exploit 3D object detections from monocular RGB frames for both, the object-based map construction, and for globally localizing in the constructed map. To tailor the approach to a target environment, we propose an efficient way of generating 3D annotations to finetune the 3D object detection model.
  We evaluate our map construction in an office building, and test our long-term localization approach on challenging sequences recorded in the same environment over nine months. The experiments suggest that our approach is suitable for constructing metric-semantic maps, and that our localization approach is robust to long-term changes. 
  Both, the mapping algorithm and the localization pipeline can run online on an onboard computer. 
  We release an open-source C++/ROS implementation of our approach.
\end{abstract}

\section{Introduction}
\label{sec:intro}

Localization and map construction are essential capabilities for mobile autonomous systems. Object-based maps, coupled with semantically-augmented localization are the foundation for more complex robotics tasks such as navigation and manipulation, as well as AR/VR applications. They enable to estimate the 3D geometry of the environment, and enrich it with semantic information.
We focus on the metric-semantic map construction coupled with long-term object-based localization, using only monocular RGB frames and a floor plan prior. 
We use RGB cameras instead of \mbox{RGB-D} due to their lower power consumption and bandwidth requirements. 
For both tasks, we are interested in approaches that operate online on a mobile platform.

Previous works on 3D mapping leverage 3D reconstruction techniques to create a geometric description of the environment~\cite{oleynikova2017iros}~\cite{palazzolo2019iros}. In recent years, the progress in semantic segmentation and object detection enabled the integration of semantic information into 3D reconstruction~\cite{grinvald2019ral}~\cite{kostavelis2015jras}~\cite{rosinol2020icra}. Some works have put emphasis on highly-accurate and detailed reconstruction of the environment~\cite{li2021ral}~\cite{runz2020cvpr}. Other approaches use 3D bounding boxes as a valuable and compact abstraction for objects~\cite{li2021iccv}~\cite{yang2019tro}. Here, we leverage 3D object detection for enriching a floor plan map with metric-semantic information suitable for long-term object-based localization. 
Floor plan maps are commonly available and store information about unchanging structures, such as walls. Localization in floor plan maps is challenging due to the sparsity of geometric information; its advantage, however, is that information stored in floor plans rarely change. Often, additional sources of information are used to support robust localization, such as textual cues~\cite{zimmerman2022iros} and WiFi signals~\cite{ito2014icra}.
3D object-based maps can be used to improve long-term localization in floor plan maps. Works such as the ones by Li~\etalcite{li2021iccv}~\cite{li2021ral} allow the construction of object-based maps from monocular RGB frames, but they do not include static structural elements such as walls, which have a critical importance for localization and planning tasks.


\begin{figure}[t]
  \centering
  \includegraphics[width=0.82\linewidth]{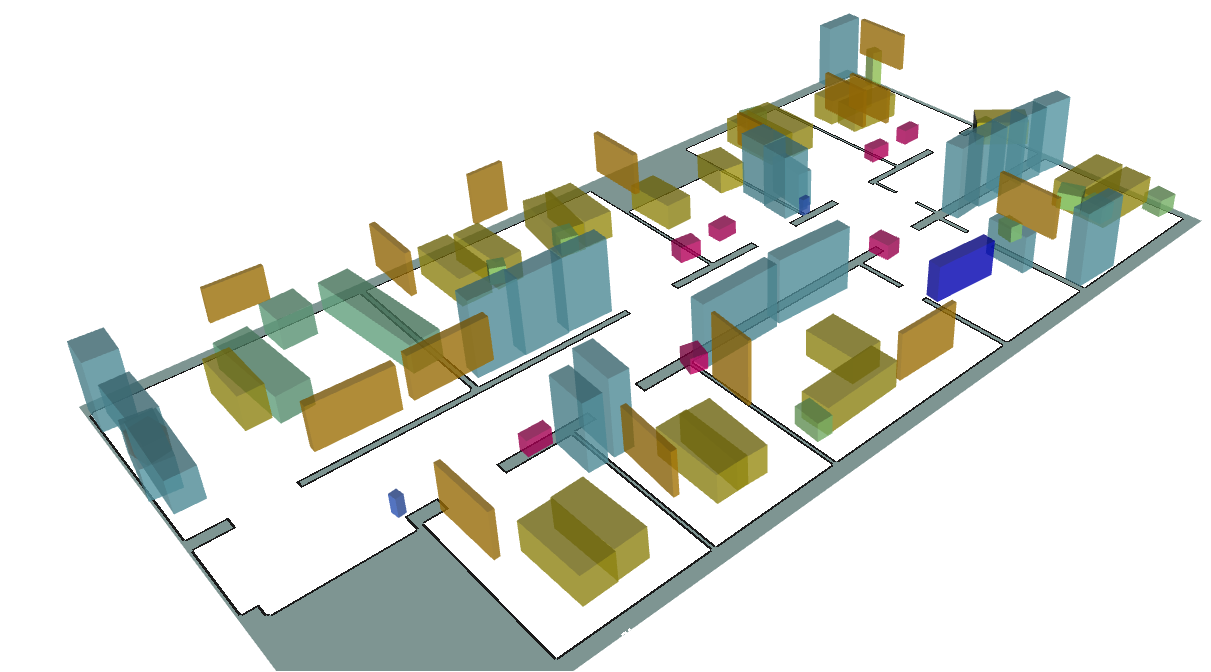}
  \caption{A 3D semantic metric map combining a floor plan with 3D object bounding boxes built using our approach. This map is used for long-term localization in dynamic indoor environments. Different box colors indicate different object classes.}
  \label{fig:motivation}
\end{figure}


The main contribution of this paper is a global localization and object-based mapping system using 3D semantic information suitable for long-term operations in dynamic environments.
We address the difficulty of creating 2D and 3D labels by proposing an efficient method for label generation from RGB frames. These labels can then be exploited to achieve accurate performance on the target environment by finetuning off-the-shelf detection models.
We analyze the performance of our detector, creating a probabilistic detection model that benefits both, map creation and object-based global localization.  
We utilize 3D object detection to construct object-centric maps, as seen in \figref{fig:motivation}, augmenting readily-available floor plan maps with semantic information. 
We provide a global localization system for the pre-built object-based map with an uncertainty-aware sensor model for 3D object information, relying solely on monocular cameras. 
In sum, our approach is able to
 (i) generate 3D labels for fine-tuning of 3D object detection models,
(ii) enrich floor plans with object information,
(iii) and localize in such maps
in an online fashion, using onboard computers.
These claims are backed up by the paper and our experimental evaluation.

\section{Related Work}
\label{sec:related}


\subsection{3D Object Detection}
Scene understanding is the ability of recognizing objects and obtaining a semantic interpretation of the surrounding environment. In the last years, deep learning enabled tremendous advancements in image-based object detection~\cite{girshick2014cvpr}~\cite{redmon2016cvpr}, semantic segmentation~\cite{seichter2021icra}, and panoptic segmentation~\cite{cheng2020cvpr}~\cite{kirillov2019cvpr-ps}~\cite{sodano2022arxiv}.
In the 3D domain, object detection approaches aimed to reproduce the efforts of 2D object detection in 3D~\cite{shi2019cvpr}~\cite{wang2021iccv}. Most approaches, however, took alternative paths. Qi~\etalcite{qi2019iccv-dhvf} propose an end-to-end 3D object detection network inspired by the generalized Hough voting. The authors also propose an extension fusing 2D and 3D voting for boosting 3D object detection~\cite{qi2020cvpr}. Recent efforts in 3D object detection tackle the problem by processing more than one frame at a time~\cite{li2021iccv}~\cite{wang2022corl}, since performances of single-view approaches~\cite{li2021ral} are degraded by the depth-scale ambiguity~\cite{li2021iccv}. Recently, Brazil~\etalcite{brazil2022arxiv} proposed a single-view approach, called Cube R-CNN, that achieves state-of-the-art results for 3D object detection and solves the depth-scale ambiguity by introducing a training objective that incorporates a virtual depth.

\subsection{Semantic Mapping}

Map construction is a crucial elements of most robotics applications, and object-based maps are a step further towards higher level scene understanding.
In recent years, learning-based approaches were applied to 3D object-based map construction. 
Many works were aimed at object-aware mapping~\cite{grinvald2019ral}~\cite{kostavelis2015jras} or simultaneous localization and mapping (SLAM)~\cite{mccormac20183dv}~\cite{stuckler2012iros} with RGB-D cameras. 
In this paper, we focus on object-based mapping with known poses using monocular RGB images. For this task, Li \etalcite{li2021iccv} use a graph neural network for data association and a super-quadratic formulation for their multi-view optimization.
Li \etalcite{li2021ral} use Bayesian filtering to track objects and the Hungarian algorithm for data association. They perform a coarse-to-fine reconstruction, first representing the objects only as localized bounding boxes, and using shape codes for a detailed reconstruction. 
A similar approach is the one introduced by Runz~\etalcite{runz2020cvpr}, that uses instead ray clustering for data association. 


\subsection{Semantic Localization}
Long-term global localization in changing environments is a challenging task. Researchers investigated different sources of information such as textual cues~\cite{cui2021iros}~\cite{zimmerman2022iros}, wifi signal strength~\cite{miyagusuku2018ral}, and semantic cues to support pose estimation. 
Rottmann~\etalcite{rottmann2005aaai} use a semantically-labeled occupancy grid map
to localize in a Monte Carlo Localization (MCL)~\cite{dellaert1999icra} framework, preforming place classification from RGB frames.
Mendez~\etalcite{mendez2018icra} extract semantic information about structural elements (walls, windows, doors) from floor plans, classify the pixels of an RGB-D image and match it against the semantic floor plan to localize. 
Both approaches do not use an explicit object-based map. 
Atanasov~\etalcite{atanasov2015ijrr} use objects as landmarks, defined by their 3D pose, semantic class and possible shape priors. The object are detected using 
a deformable part model~\cite{felzenszwalb2013acm}, and the semantic information is integrated into a MCL framework. 
Ranganathan~\etalcite{ranganathan2007rss} suggest a different representation for a semantic map, by using a constellation model. They detect objects using hand-crafted features, and rely on depth information provided by stereo cameras. 
Yi \etalcite{yi2009iros} localize using a topological graph, where each node is characterized by the semantic objects in its vicinity. 
Similarly to these approaches, we incorporate the prediction uncertainty into our localization approach and use sparse object-based maps. We provide a pipeline for aquiring the metric semantic map. Most other approaches use 2D object detection, while we utilize 3D object information.
Zimmerman~\etalcite{zimmerman2023ral} incorporate high and low-level semantic cues from 2D object detection and geometric information from 2D LiDAR scanners, but use manual map creation. 
In contrast to this, our approach does not require LiDAR input. 

\begin{figure}[t]
  \centering
  \includegraphics[width=0.9\linewidth]{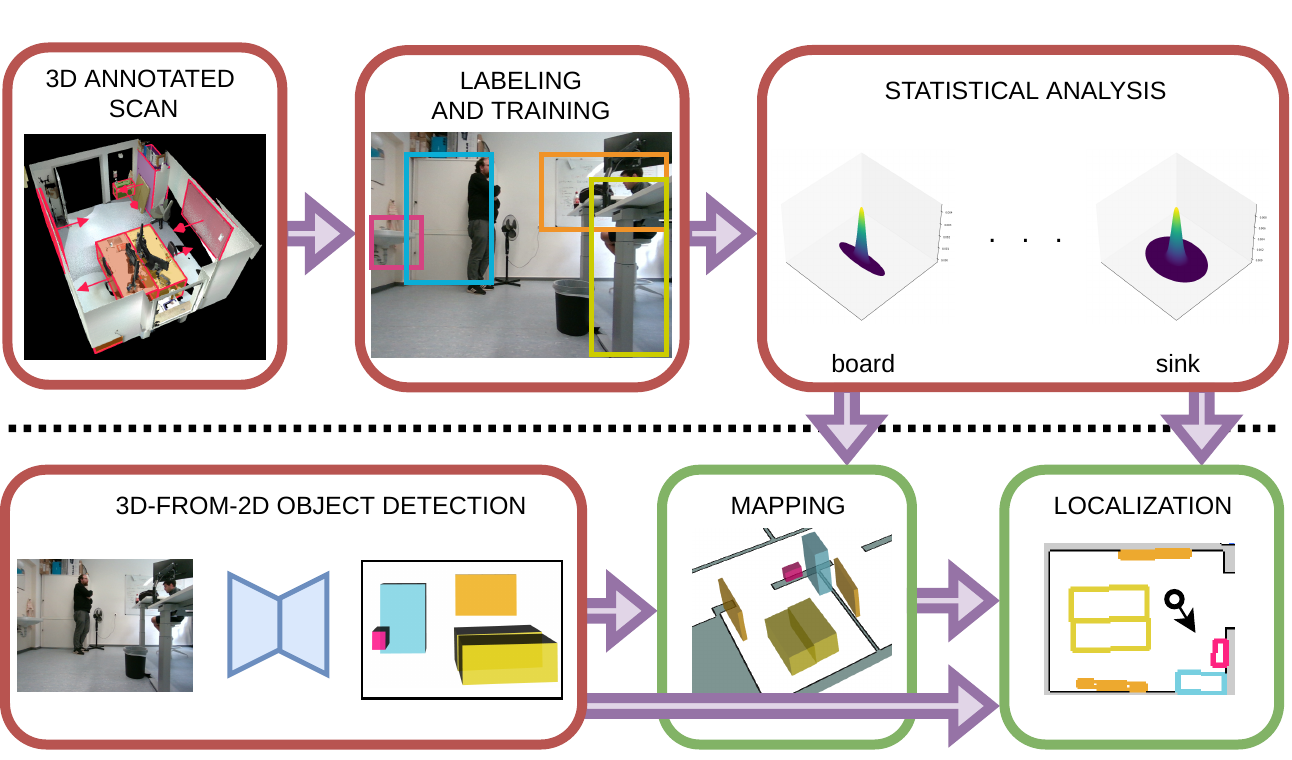}
  \caption{An overview of our approach. Top row: offline pre-computation to adapt the approach to a specific environment. Bottom row: 3D-from-2D object detection and mapping, that can be executed on demand when the environment undergoes structural changes and a map update is necessary.}
  \label{fig:overview} 
\end{figure} 

\section{Our Approach}
\label{sec:main}

We aim to enhance sparse floor plans 
with semantic cues and globally localize in these object-based maps using monocular cameras.
In Sec.~\ref{sec:labeling}, we present a way of creating 3D labels for fine-tuning 3D object detection models. 
Using such labels, we show how to fine-tune an object detection model and learn a noise model, see Sec.~\ref{sec:stat}, and use it to build a probabilistic object-based map.
We construct a standard metric-semantic map using 3D object detections on posed RGB images, detailed in Sec.~\ref{sec:map}. 
We exploit the object-based maps for object-based global localization, as described in Sec.~\ref{sec:localize}. An overview of our approach is visualized in \figref{fig:overview}.

\begin{figure}[t]
  \centering
  \includegraphics[width=0.95\linewidth]{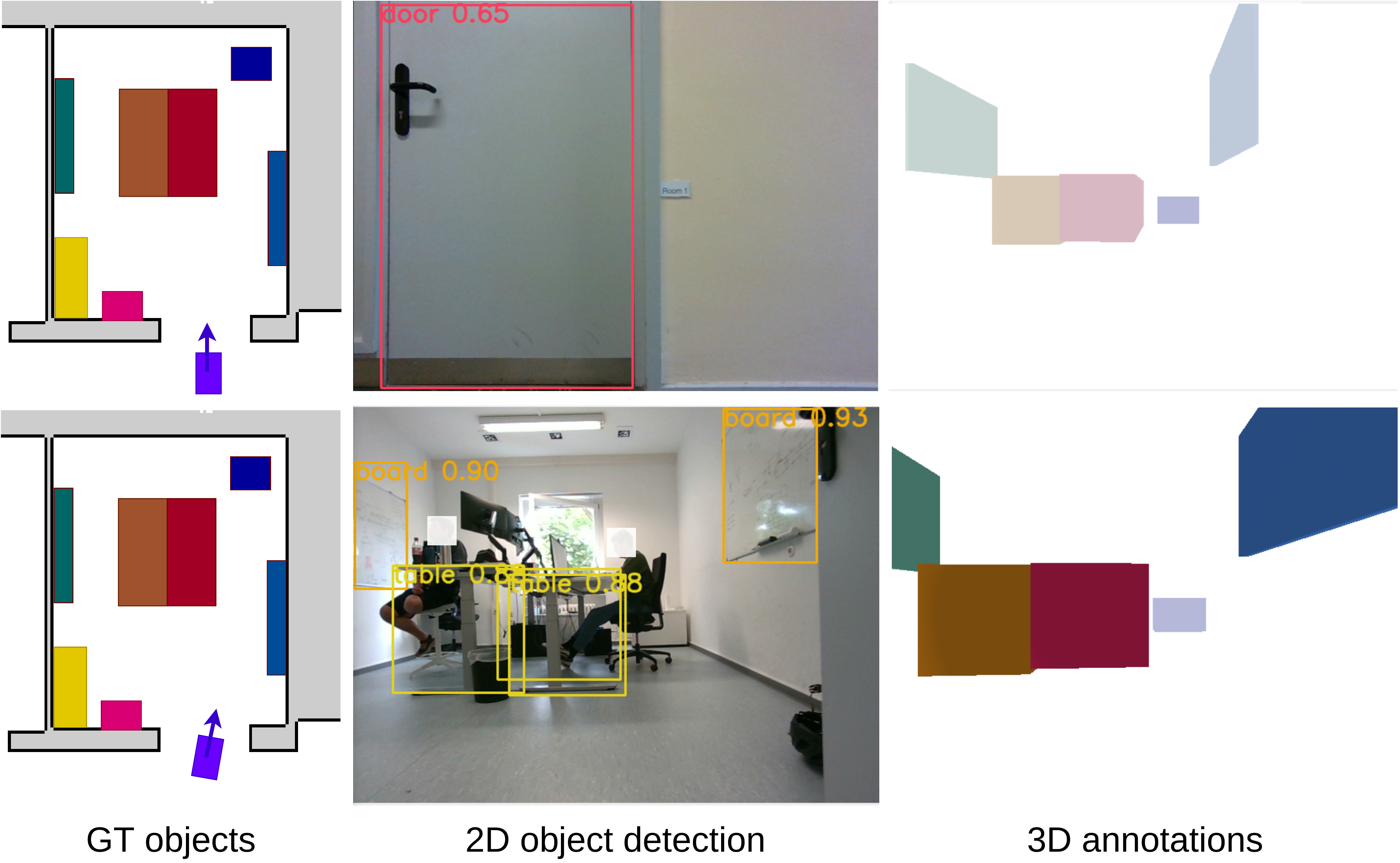}
  \caption{2D object detection to create better 3D annotations. Top: when rendering the ground truth objects
  from the camera, we have no information about dynamic objects like closed doors, which results in wrong annotations. The 2D object detection detects a door, and none of the objects in the 3D map. Therefore, no 3D annotations 
  are generated (faded colors). Bottom: the 2D object detection detects tables and boards, but not the drawers due to occlusion. Therefore 3D annotations are generated only for the boards and the tables. }
  \label{fig:rendering}
\end{figure}

\subsection{Label Generation for 3D Object Detection}\label{sec:labeling}
The performance of off-the-shelf 3D object detection models are often not accurate enough for the purpose of map building and they possibly focus on classes that are not beneficial for the purpose of indoor localization. For this reason, mapping and localization systems might need to finetune an existing model to the environment of interest. 
3D object detection from RGB images requires 3D bounding boxes annotations that include the object dimension, rotation and translation relative to the camera frame, in addition to 2D bounding box annotation for the RGB image with semantic category. 
Often, truncation and visibility of objects are also needed. Truncation refers to the percentage of the object in the camera frustum, while visibility is a measure of occlusion of an object by other object in the scene.  Providing accurate labels in a real-world environment is challenging. 
In our approach, we construct the 3D labels based on a 3D model of the environment extracted from a 3D scan, 2D object detector and posed RGB images. 
As a pre-processing step, we manually annotate the 3D model of the environment with 3D bounding boxes of objects of interest. 
We project the 3D bounding boxes onto the posed camera frame using
\begin{align}
  \q{x} = \mq{K} \, \begin{bmatrix}
    \b{R} & \b{t}\\
    0 & 1\\
  \end{bmatrix} \, \q{X},
  \label{eq:reprojection}
\end{align}
where $\q{X}=(x, y, z, 1)^\top$ is a 3D point in the world coordinate system in homogenous coordinates, 
$\mq{K} \in \mathbb{R}^{3 \times 3}$ is the camera calibration matrix, $\b{R}$ and $\b{t}$ are the camera rotation matrix 
and translation vector, and $\q{x}$ is a point in the image plane in homogenous coordinates.
We then determine the visibility and truncation for every annotated object.
To determine if an object is occluded by a dynamic obstacle, such as a person or a closed door, or by a static obstacle that is not annotated, we use a 2D object detector, as seen in \figref{fig:rendering}. The detector is trained on classes of interest including dynamics and is made available by Zimmerman~\etalcite{zimmerman2023ral}.
For every object detected using the 2D object detector, we match a previously-annotated 3D bounding box based on the semantic class and the IoU between the projected and the detected 2D bounding box.
In this way, for every posed RGB frame, the system annotates the detectable 3D objects, including their relative pose in the camera frame, dimensions, semantic class, 2D bounding boxes, visibility and truncation.
Using these labels, we fine-tune a 3D object detector. We choose Cube R-CNN~\cite{brazil2022arxiv}, but our implementation allows integration of any other 3D object detector with the same output structure. Further details about the architecture and training procedure can be found in \secref{sec:exp}.

\begin{figure}[t]
  \begin{subfigure}{0.15\textwidth}
    \includegraphics[width=0.99\linewidth]{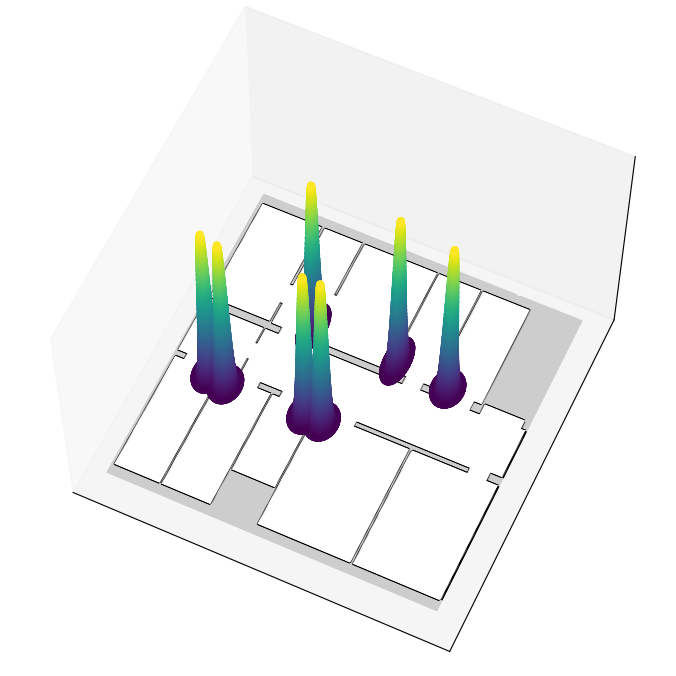}
    \caption{Sinks}
  \end{subfigure}
  \begin{subfigure}{0.15\textwidth}
    \centering
    \includegraphics[width=0.99\linewidth]{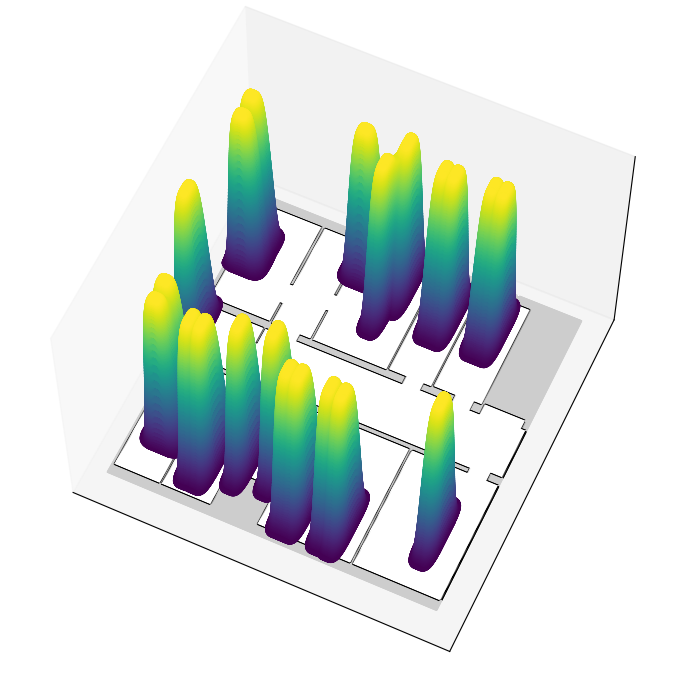}
    \caption{Tables}
  \end{subfigure}
  \begin{subfigure}{0.15\textwidth}
    \centering
    \includegraphics[width=0.99\linewidth]{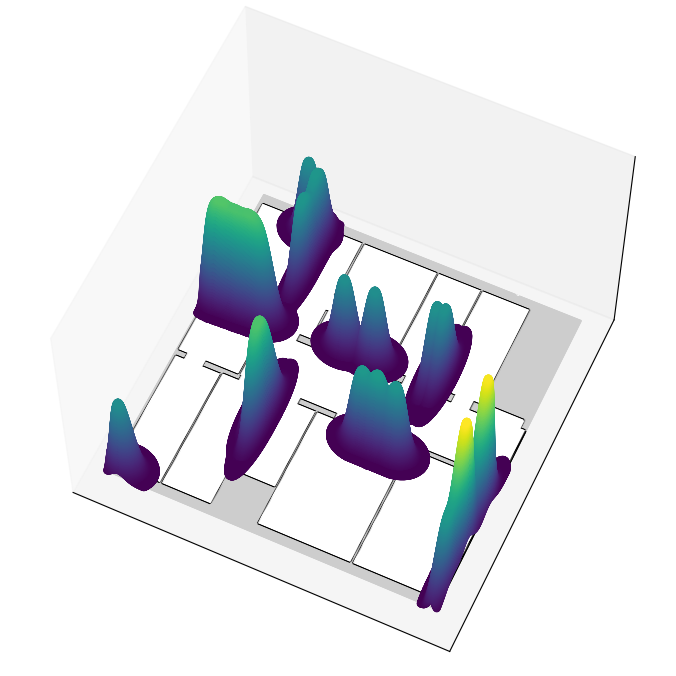}
    \caption{Cabinets}
  \end{subfigure}
  \centering
  \caption{Object probability map $m_p$ which contains the per-object distribution $p_o(c \mid l)$, for specific classes of interest.}
  \label{fig:objprobmap}
\end{figure}

\subsection{Statistical Analysis of 3D Object Detections}\label{sec:stat}
Given posed RGB frames, capturing objects belonging to classes of interest,
we run inference on the images using Cube R-CNN, fine-tuned with our labels.
Given two non-overlapping object predictions of the same class, a key issue is determining whether they belong to the same instance suffering from noisy detection or they are two separate instances in close vicinity. This is particularly common with semantic classes whose bounding boxes have one dimension which is substantially smaller than the other, such as whiteboards. To address this problem, we analyze the per-class characteristic noise for detections. 
For each class, we build a 2D probability distribution by matching the 3D predictions to map objects~$\mathbb{O}$, 
and projecting them in 2D on the ground plane. 
For matching, a predicted object can be assigned to a ground truth object only if the euclidean distance between the center of the predicted object $\b{c}_o$ and the center of the ground truth object $\b{c}_{GT}$ is smaller than a threshold $\delta$.

In the case a predicted object can be assigned to multiple ground truth objects, we select the one with the highest 3D IoU. In case the prediction has no overlap with any ground truth, we match based on center distance only. Additionally, if no ground truth center is within distance~$\delta$ from the center of the predicted object, we discard the prediction. Then, we aggregate predictions of different objects of the same class by transforming them into an object-centric coordinate system where the center of the associated ground truth object is the origin. 
For every class, we take the matched predictions and project their center on the 2D  plane. Additionally, we discretize the 2D plane into cells of $5 \, cm$ creating a histogram of  occurrences. Then, we fit a Gaussian distribution
\begin{align}
  p(\b{c} \mid l) &= \dfrac{1}{\sqrt{2 \pi \abs{\b{\Sigma}}}} \exp{ \Big\{ -\dfrac{(\b{c} - \b{\mu})^\top \b{\Sigma}^{-1}(\b{c} - \b{\mu})}{2} \Big\} }, 
\label{eq:perclassdist}
\end{align}
where~$\b{c}$ indicates the 2D center coordinate of the predicted object, and $l$ is its semantic class.

Then, we transform the per-class distribution~$p(\b{c} \mid l)$ to be centered around each map object~$o$ of class~$l$. 
We shift the mean~$\b{\mu}$ and rotate the covariance~$\b{\Sigma}$ of the class-specific distribution in local frame, according to the projected 2D center point~$\b{c}_o$ and the rotation matrix~$\b{R}_o$ of an object:
\begin{equation}
  \b{\mu}_o = \b{\mu} + \b{c}_o, \qquad
  \b{\Sigma}_o = \b{R}_o \, \b{\Sigma} \, \b{R}_o^\top.
\label{eq:shiftdist}
\end{equation}
Thus, the per-object distribution $p_o(\b{c} \mid l)$ is given by the parameters $\b{\mu}_o$ and $\b{\Sigma}_o$ of~\eqref{eq:shiftdist}.

We build the object probability map $m_p$ by composing the individual Gaussians of the map objects $p_o(\b{c} \mid l), \forall o \in \mathbb{O}$.
Our object probability map $m_p$ can be seen in \figref{fig:objprobmap}.

\subsection{3D Semantic Map Construction}\label{sec:map}
Given posed RGB images and a 3D object detector,
we construct a metric semantic map to enhance a floor plan map.
We define an object as 
\begin{equation}
o = \{\b{c}, \, \b{D}, \,\b{R}, \, \mathrm{I}_{\mathrm{active}}, \, n_{\mathrm{skip}}, \, n_{\mathrm{match}} \},
\label{eq:objdef}
\end{equation}
where $\b{c}$ is the center of the object, $\b{D} = (W, H, L)$ is the object's bounding box dimensions, $\b{R}$ is the orientation of the bounding box, 
$\mathrm{I}_{\mathrm{active}} \in \{0, \, 1\}$ is a state that indicates whether an object is active or not, $n_{\mathrm{skip}}$ and $n_{\mathrm{match}}$ are two object-specific counters that will be explained in the following.
An object is in the active state if it is in the camera frustum and is not occluded by other objects. We test for the visibility of an object by rendering the 3D scene into the camera frame, constraining the visibility with walls extracted from the floor plan map.
We aggregate consecutive detections into a short-term, local map $\hat{m}_g$ by associating detected objects across different frames. Active objects are associated by means of the Hungarian algorithm~\cite{kuhn1955nrlq} and a cost function defined by
\begin{align}
  \mathcal{C}(o_1, \, o_2) &= \dfrac{1}{2} \, \big(\mathcal{C}_{\mathrm{IoU}} + \mathcal{C}_{\mathrm{cen}} \big)\\
  \mathcal{C}_{\mathrm{IoU}} &= 1 - \mathrm{IoU}(o_1, \, o_2) \\
  \mathcal{C}_{\mathrm{cen}} &= 1 - p_{o_1}(\b{c}_{o_2} \mid l_{o_2}),
  \label{eq:costfunc}
\end{align}
where $o_1$ and $o_2$ are detected objects, $p_{o_1}(\b{c}_{o_2} \mid l_{o_2})$ represents the goodness score of a center prediction based on the statistical analysis in Sec.~\ref{sec:stat}, $\b{c}_{o_2}$ and $l_{o_2}$ are the center and the semantic class of object $o_2$, respectively.
After the robot has moved more than $d_{xy}$ or rotated more than $d_\theta$, we integrate the objects of $\hat{m}_g$ into the global object map $m_g$ using the matching strategy described above.
Given the associations computed by the Hungarian algorithm, we merge matched objects if the cost in~\eqref{eq:costfunc} is below a threshold $\tau_{\mathrm{cost}}$. Otherwise, we initialize a new map object. 
When objects are merged, we increase the $n_{\mathrm{match}}$ count. If an active map object was not associated with a prediction, we increase the $n_{\mathrm{skip}}$.
When merging a prediction $o_1$ to a map object $o_2$, we use a weighted average to update the center and the dimensions of the bounding box:
\begin{equation}
  \begin{split}
  \b{c}_{o_2} & = \dfrac{n_{\mathrm{match}}^{o_2} \, \b{c}_{o_2} + n_{\mathrm{match}}^{o_1} \, \b{c}_{o_1}}{n_{\mathrm{match}}^{o_1} + n_{\mathrm{match}}^{o_2}} \\
  \b{D}_{o_2} & = \dfrac{n_{\mathrm{match}}^{o_2} \, \b{D}_{o_2} + n_{\mathrm{match}}^{o_1} \, \b{D}_{o_1}}{n_{\mathrm{match}}^{o_1} + n_{\mathrm{match}}^{o_2}}.
  \end{split}
\end{equation}

We update the rotation matrix by computing the weighted average and extracting the rotation matrix via SVD as proposed by Moakher~\etalcite{moakher200siam}:
\begin{align}
    \b{U} \, \b{\Sigma} \, \b{V}^\top &= \mathrm{SVD} \bigg( \dfrac{n_{\mathrm{match}}^{o_2} \, \b{R}_{o_2} + n_{\mathrm{match}}^{o_1} \, \b{R}_{o_1}}{n_{\mathrm{match}}^{o_1} + n_{\mathrm{match}}^{o_2}} \bigg) \\ 
    \b{R}_{o_2} &= \b{U} \, \b{V}^\top.
\end{align}

Additionally, we update the weight of the object by summing up $n_{\mathrm{match}}$.
We purge objects from the map when 
\begin{equation*}
  \dfrac{n_{\mathrm{match}}}{n_{\mathrm{skip}}} < \tau_{\mathrm{purge}},
\end{equation*}
where $\tau_{\mathrm{purge}}$ is an empirically-chosen threshold.

We obtain room segmentation by applying morphological operations and connected-component analysis on the floor plan. This allows us to associate objects to rooms, so we update only objects located in the room we are currently in.



\subsection{3D Semantic Localization}\label{sec:localize}
We globally localize using an MCL~\cite{dellaert1999icra} framework. MCL is a particle-based approach for estimating
the robot state~$\b{x}_t$, given the map~$m$, odometry input~$\b{u}_t$ and the observations~$\b{z}_t$. Each particle~$\b{s}_t$ is represented
by a state~$\b{x}_t$ and a weight~$w_t$, and the purpose of the sensor model is to assign a weight~$w_t$ to a particle~$\b{s}_t$ given an observation~$\b{z}_t$ and the map~$m$. In our case, we localize in a 2D occupancy grid map, and therefore the robot state is~$\b{x}_t = (x, \, y, \, \theta)^\top$.

Our sensor model is based on the probabilistic analysis of the accuracy of the 3D object detection model, see Sec.~\ref{sec:stat}. 
Our observation~$\b{z}$ includes the object class~$l$, the confidence score~$f$ and the object 3D bounding box~$\b{b}_{\mathrm{3D}}$ in the coordinate frame of the camera.
For every particle~$\b{s}_t$, we transform the prediction into world frame using the particle state~$\b{x}_t$, and we project it on the ground, obtaining the corresponding 2D bounding box~$\hat{\b{b}}_{\mathrm{2D}}$.

To compute the weight of every particle we consider two measures -- the object-based likelihood~$p_o(\b{z} \mid m_p, \b{x}_t)$ and the shape similarity score~$p_s$.
Using the object probability map~$m_p$ computed in Sec.~\ref{sec:stat}, we sample~$p_o$ at the location indicated by the center~$\b{c}$ of the transformed 2D bounding box. 
To compute the likelihood of predicting a center~$\b{c}$ of class~$l$, we consider the corresponding class distribution for every object given by the object probability map~$m_p$:
\begin{align}
  p_o(\b{z} \mid m_p, \, \b{x}_t) = \max_{ \{ o\in \mathbb{O} \, | \, l_o = l \} } p_o(\b{c} \mid l),
 \label{eq:globaldist}
\end{align}
where~$l_o$ is the semantic class of object~$o$. For data association, we match the prediction to map object with the highest likelihood, referred to as~$o_{\mathrm{max}}$.

The metric semantic map~$m_s$, constructed in Sec.~\ref{sec:map}, stores the 3D bounding box of~$o_{\mathrm{max}}$, which we project to 2D bounding box~$\hat{\b{b}}^{\mathrm{max}}_{\mathrm{2D}}$.
The shape similarity score is computed from the IoU of the 2D bounding boxes~$\hat{\b{b}}_{\mathrm{2D}}$ and~$\hat{\b{b}}^{\mathrm{max}}_{\mathrm{2D}}$:
\begin{align}
  p_g(\b{z} \mid m_g, \b{x}_t) &= \exp \bigg(\!\! - \!\bigg( 1 \!-\! \mathrm{IoU} \bigg( \hat{\b{b}}_{\mathrm{2D}}, \, \hat{\b{b}}^{\mathrm{max}}_{\mathrm{2D}} \bigg) \bigg) \bigg).
  \label{eq:iouscore}
 \end{align}

 The probability of detecting an object from particle state~$\b{x}_t$ is given by
 \begin{align}
  p(\b{z} \mid m, \, \b{x}_t) = p_o \, p_g + (1 - p_o) \, \eta,
  \label{eq:combined}
\end{align} 
where~$\eta$ is an empirically computed constant, representing the weight of a false data association, and~$p_o$ and~$p_g$ are defined in Eqs.~\ref{eq:globaldist} and~\ref{eq:iouscore}, respectively. 
When~$K$ objects are detected in a single frame, we compute the overall particle weight as a geometric average 
\begin{align} 
  p(\b{z}_t \mid m, \, \b{x}_t)  = \prod_{k=1}^K{ p(\b{z}_t^k \mid m, \, \b{x}_t) }^{\frac{1}{K}}.
  \label{eq:singw}
 \end{align}



\begin{table}[t]
  \caption{Algorithm parameters} 
   \centering
   \resizebox{\columnwidth}{!}{
 \begin{tabular}{ccccccc}\toprule
 Method & $\sigma_{\mathrm{odom}}$  &  $\sigma_{\mathrm{obs}}$ & $r_{\mathrm{max}}$ & $ \tau_{s} $ &  $d_{\mathrm{xy}}$ & $d_{\theta}$\\ \midrule
 Ours & (0.15\,m, 0.15\,m, 0.15\,rad) & - & - & - &   0.1\,m & 0.03\,rad\\
 HSMCL & (0.15\,m, 0.15\,m, 0.15\,rad) & 6.0 & 15.0\,m & 0.6  & 0.1\,m & 0.03\,rad\\

  \bottomrule
 \end{tabular}
 }
 \label{tab:parameters}
\end{table}

\section{Experimental Evaluation}
\label{sec:exp}

%
%
We present our experiments to show the capabilities of our method. The results of our experiments also support our key claims, which are: our approach is able to
(i) generate 3D labels for fine-tuning of 3D object detection models,
(ii) enrich floor plans with object information,
(iii) and localize in such maps
in an online fashion, using onboard computers.

\begin{table*}[t] 
  \caption{Computed metrics for the two constructed maps compared to the  map obtained with a FARO 3D scan. KP was constructed based on known poses from 
  infrastructure, and ICP was constructed with poses extracted from 2D LiDAR ICP.}
  \centering 
  \begin{tabular}{cccccccccccccc} 
    \toprule
    & map & board & cabinet & desk & drawers & fire ext. & oven & plant & sink & sofa & table & AVG \\
    \midrule
    IoU & KP & 0.54 & 0.75 & 0.79 & 0.39 & 0.55 & 0.85 & 0.61 & 0.77 & 0.68 & 0.76 & 0.67 \\
    Pr & KP & 1.00& 1.00& 1.00& 0.56& 1.00& 1.00& 1.00& 1.00& 1.00& 1.00& 0.96\\
    Rc & KP & 0.94& 0.92& 1.00& 0.62& 1.00& 1.00& 1.00& 1.00& 1.00& 0.95& 0.94\\
  \midrule
  IoU & ICP&  0.51 & 0.70 & 0.77 & 0.40 & 0.66 & 0.68 & 0.60 & 0.65 & 0.77 & 0.75 & 0.65 \\
  Pr & ICP & 1.00& 1.00& 1.00& 0.56& 1.00& 1.00& 1.00& 1.00& 1.00& 1.00& 0.96\\
  Rc & ICP & 1.00& 0.92& 1.00& 0.62& 1.00& 1.00& 1.00& 1.00& 1.00& 0.95& 0.95\\
  \bottomrule 
  
\end{tabular}
\label{tab:mapping_eval} 
\end{table*}

\subsection{Experimental Setup}





To assess the performance of our approach, we made multiple recordings in our building.
Our data collection platform was a Kuka YouBot with 2 Hokuyo UTM-30LX LiDARs, 
wheel encoders, 4 cameras with a joint coverage of $360^{\circ}$ field-of-view, and an up-looking camera used strictly for evaluation purposes. 
The recordings span across 9 months, capturing changes to the lab furniture, varying amount of clutter, human movement
and opening and closing of doors.

To extract precise ground truth information about the robot's pose, we use an external localization infrastructure based on densely placed (approx. $1 \, \mathrm{tag/m}^2$) AprilTags, covering the ceiling of each room and corridor of our lab. In every frame captured with the up-looking camera, we detect multiple AprilTags computing the pose estimation in a least-squares fashion and achieving accuracy of under $3 \, \mathrm{cm}$.
A high resolution point cloud of the lab, generated with a terrestrial laser scanner, was also used to produce the 3D labels used in \ref{sec:labeling}, and the localization 
infrastructure was used to generate the poses for the RGB frames used the metric-semantic mapping in \ref{sec:map}.

For Cube R-CNN~\cite{brazil2022arxiv}, we fine-tuned the ResNet34-based model~\cite{he2016cvpr} the authors provide for indoor perception. For doing this, we created a list of objects of interest suitable for long-term localization. The success of our approach depends on the appropriate choice of the classes of interest. It is important to consider the stability of the object classes, as discussed in~\etalcite{zimmerman2023ral}, and their observability and map coverage. We optimize our model with stochastic gradient descent for 100 epochs, with an initial learning rate of 0.0015 and a batch size of 12. For training, we recorded sequences T1-T7 on the second floor of our building.

As baseline, we compare against HSMCL~\cite{zimmerman2023ral}, another semantic MCL framework. 
All experiments were executed with $5000$ particles. The parameters for the different approaches are reported in \tabref{tab:parameters}. 
We consider three metrics, the success rate, absolute trajectory error~(ATE) after convergence and convergence time. 
We define convergence as the time when the estimate pose is within $0.3 \, m$ radius of the ground truth pose, and the orientation is within 
$\frac{\pi}{4}$. After globally localizing, the tracked pose must not diverge for an accumulated $1.5 \, s$.
A localization run is successful if convergence is achieved in the first 95\% of the sequence time and the tracked pose 
does not diverge. Each sequence was evaluated multiple times to account for the inherent stochasticity of the MCL framework, and 
the success rate is computed over multiple runs. 

\begin{figure}[t]
  \centering
  \includegraphics[angle=90,width=0.32\linewidth]{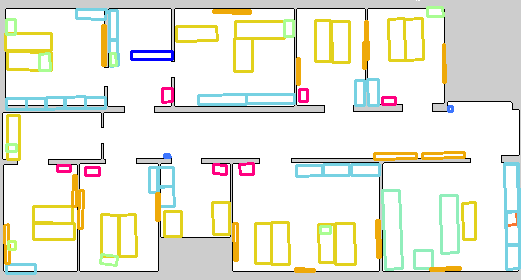} 
  \includegraphics[angle=90,width=0.32\linewidth]{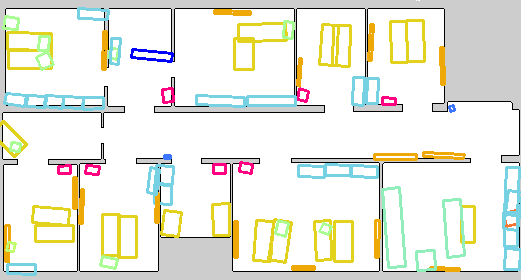} 
  \includegraphics[angle=90,width=0.32\linewidth]{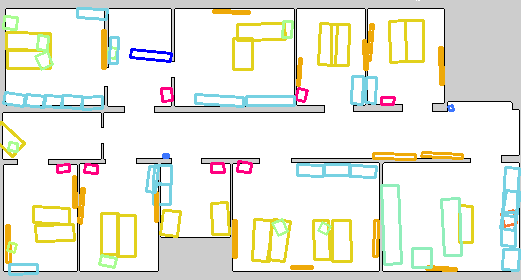}
  \caption{2D projection for the 3 maps. Left: ground truth map obtained with terrestrial laser scanner. Middle: KP map. Right: map built using scan matching (ICP).}
  \label{fig:maps}
\end{figure}
\subsection{Mapping} 
To support our second claim, we evaluate the quality of our mapping pipeline. We use sequence M1, which include ground truth poses, RGB stream, and 2D LiDAR scans, while the robot was traversing the entire second floor.
We evaluate our constructed maps by matching them against a ground truth (GT) map, extracted with a highly accurate 3D terrestrial laser scanner, and annotated manually. The creation of the ground truth map is labour-intense offline procedure that requires specialized sensors and equipment, providing sub-centimeters accuracy. It has been carried out for evaluation purposes only. 
Our constructed maps are built using the metric-semantic mapping pipeline describe in Sec. \ref{sec:map}, using data for sequence M1. 
For the first constructed map, we used precise poses extracted from our localization infrastructure, and we refer to it as known poses (KP) map.
We also generated a map using sensor-based poses, estimated via ICP on 2D LiDAR scans. We aligned the estimated poses to an existing grid map, such as floor plan, by initializing the ICP with a known pose. We believe the poses could also be extracted with low-drift visual-inertial odometry pipelines~\cite{rosinol2020icra}.
We evaluated the accuracy of the generated maps directly by comparing it to the ground truth map, considering the IoU between objects in the object-based map, and precision and recall. 
As can be seen in \tabref{tab:mapping_eval} and \figref{fig:maps}, the constructed maps are well-matched to the ground truth map:
the IoU scores between the ground truth map and our generated map are high, while precision and recall are close to $1$, indicating a low number of false positives and false negatives. 
We also evaluate the performance of robot localization based on the constructed maps. The localization accuracy for 
the KP map and the ICP map is on par with the accuracy for the ground truth (GT) map, see \tabref{tab:success_eval}, suggesting that our maps are well-suited for localization.

\subsection{Long-Term Localization in CAD Floor Plans}
To assess the performance of our localization approach, we recorded sequences R1-R10, through our building, spanning across nine months. These sequences include the addition and removal of furniture, dynamic obstacles, 
and quasi-static changes such as the closing and opening of doors. The starting points for the localization sequences were spread between different rooms and the corridor. 
Accuracy and success rate are reported in~\tabref{tab:success_eval} for the three different maps. Each map was generated once, and has not been updated during the evaluation period. Convergence time averaged over sequences R1-R10 is $10.9 \, s$ for ground truth map and $12.1 \, s$ for KP map. 





\begin{table*}[t] 
  \caption{Evaluation of the map construction quality through long-term localization performance. The success rate for all maps on all reported sequences is $100 \%$. We report ATE in $[\mathrm{rad}/m]$ format for 10 sequences recorded all across our lab in the span of nine months.}
  \centering 
  \resizebox{\textwidth}{!}{
\begin{tabular}{ccccccccccccc}
  \toprule
  Method     & Map   & R1 & R2 & R3 & R4 & R5 & R6 & R7 & R8 & R9 & R10 & AVG\\  \midrule
  Ours &GT  & 0.089/0.14        & 0.065/0.11        & 0.070/0.14        & 0.043/0.17        & 0.065/0.16        & 0.075/0.15        & 0.073/0.17        & 0.077/0.20        & 0.060/0.17        & 0.047/0.12        & 0.066/0.15 \\
  Ours &KP  & 0.082/0.14        & 0.098/0.16        & 0.082/0.15        & 0.041/0.14        & 0.088/0.22        & 0.075/0.14        & 0.073/0.16        & 0.085/0.23        & 0.065/0.15        & 0.042/0.14        & 0.073/0.16\\
  Ours & ICP & 0.078/0.14        & 0.115/0.11        & 0.085/0.16        & 0.039/0.15        & 0.077/0.18        & 0.081/0.14        & 0.074/0.18        & 0.083/0.27        & 0.057/0.12        & 0.045/0.11        & 0.073/0.16 \\
  \bottomrule
\end{tabular}}
\label{tab:success_eval} 
\end{table*}

\begin{table*}[t] 
  \caption{Baseline comparison for long-term localization on the ground truth map. We report success rate and ATE in $[\mathrm{rad}/m]$ format for 10 sequences recorded all across our lab in the span of nine months.}
  \centering 
  \resizebox{\textwidth}{!}{
\begin{tabular}{ccccccccccccc}
  \toprule
  
  Method     & Map   & R1 & R2 & R3 & R4 & R5 & R6 & R7 & R8 & R9 & R10 & AVG\\  \midrule
  HSMCL	& GT & 0\% 	& 60\% 	& 100\% 	& 100\% 	& 100\% 	& 0\% 	& 100\% 	& 100\% 	& 100\% 	& 20\% 	& 68\% \\
  EDT-MCL  & GT &   100\% 	& 80\% 	& 100\% 	& 100\% 	& 100\% 	& 100\% 	& 100\% 	& 100\% 	& 100\% 	& 0\% 	& 88\% \\
  D-MCL  & GT & 100\% 	& 60\% 	& 100\% 	& 100\% 	& 100\% 	& 100\% 	& 100\% 	& 100\% 	& 80\% 	& 0\% 	& 84\% \\
  O-MCL  & GT& 100\% 	& 80\% 	& 100\% 	& 100\% 	& 40\% 	& 0\% 	& 100\% 	& 100\% 	& 100\% 	& 0\% 	& 72\% \\
  Ours &GT  & 100\% & 100\% & 100\% & 100\% & 100\% & 100\% & 100\% & 100\%  & 100\%& 100\%&100\%\\
  \midrule 
  HSMCL	&GT 	& {-}/{-}   	& {-}/{-}   	& 0.101/0.30		& 0.147/0.32		& 0.084/0.25		& {-}/{-}   	& 0.052/0.19		& 0.063/0.24		& 0.078/0.30		& {-}/{-}   	& 0.088/0.27 \\
  EDT-MCL  & GT &  0.094/0.17		& {-}/{-}   	& 0.075/0.18		& 0.043/0.20		& 0.122/0.16		& 0.076/0.20		& 0.073/0.25		& 0.058/0.23		& 0.060/0.16		& {-}/{-}   	& 0.075/0.19 \\
  D-MCL  & GT &   0.068/0.13		& {-}/{-}   	& 0.055/0.18		& 0.036/0.17		& 0.075/0.16		& 0.037/0.21		& 0.046/0.24		& 0.039/0.22		& {-}/{-}   	& {-}/{-}   	& 0.051/0.19 \\
  O-MCL  & GT& 0.082/0.14		& {-}/{-}   	& 0.054/0.15		& 0.045/0.16		& {-}/{-}   	& {-}/{-}   	& 0.056/0.18		& 0.049/0.19		& 0.061/0.14		& {-}/{-}   	& 0.058/0.16 \\
  Ours &GT  & 0.089/0.14        & 0.065/0.11        & 0.070/0.14        & 0.043/0.17        & 0.065/0.16        & 0.075/0.15        & 0.073/0.17        & 0.077/0.20        & 0.060/0.17        & 0.047/0.12        & 0.066/0.15 \\
  \bottomrule
\end{tabular}}
\label{tab:ablation} 
\end{table*}

\subsection{Baseline Comparisons for Semantic Localization}

We compare our approach to three strategies for integrating 3D object information into MCL. 
The first baseline, called EDT-MCL, extends the commonly-used beam-end point model. 
Based on the semantic metric map $m_s$ (see \secref{sec:map}), we create an Euclidean distance transform~(EDT) for each semantic class.

The second approach, D-MCL, computes the particle weight based on the object probability map $m_p$ and the likelihood $p_o$ defined in \eqref{eq:globaldist}.
The third approach, O-MCL, is based on the overlap between a predicted bounding box and the semantic-metric map $m_s$. The weight is computed based on the overlap score described in \eqref{eq:iouscore}.

As can be seen in \tabref{tab:ablation}, our approach outperforms the baselines. 
Both EDT-MCL and O-MCL do not incorporate information about the model noise. D-MCL makes use of the learned statistical information about the model performance, but only considers the center of the prediction, discarding valuable information about the object dimensions and rotation.
Our approach leverages both the geometric information from the 3D bounding box and takes into account the characteristic model noise, which results in improved performance. Unlike HSMCL, our approach does not use LiDAR information, yet the localization is more robust. 



 
\subsection{Runtime} 
Our mapping and localization approaches both run online, onboard of a mobile platform. 
On our robots, we use an Intel NUC10i7FNK and a NVidia Jetson Xavier AGX. The 3D object detection runs at $\sim$$9$ FPS on the NVidia Jetson, and the sensor model executes at $60$ FPS on the NUC10i7FNK. This performance
is sufficient to construct a 3D metric semantic map and globally localize on our mobile platform. A longer video, including live demoes for both mapping and localization, and the code can be found on our GitHub repository at {\small\texttt{https://github.com/PRBonn/SIMP}}.







\section{Conclusion}
\label{sec:conclusion}

In this paper, we presented a novel approach for semantic global localization with a complementary 3D mapping procedure to build the object-based maps used for localization. 
The augmentation of floor plan maps with the 3D metric semantic maps assists navigation in cluttered and dynamic indoor environments. 
We show that our semantically-guided localization is reliable and accurate on both, the ground truth map and the map acquired with 
our proposed pipeline, benchmarking it over a dataset of challenging scenarios spanning over nine months.  
We compared our approach to similar approaches, and the experiments show how long-term localization can benefit from 3D metric semantic maps.



\bibliographystyle{plain_abbrv}

\bibliography{glorified,new}

\end{document}